\documentclass[runningheads]{llncs}
\usepackage{graphicx}
\usepackage{amsmath}
\usepackage{amssymb}
\usepackage{wrapfig, booktabs}
\usepackage{tabularx}
\usepackage[pagebackref,breaklinks,colorlinks]{hyperref}

\usepackage[capitalize]{cleveref}
\crefname{section}{Sec.}{Secs.}
\Crefname{section}{Section}{Sections}
\Crefname{table}{Table}{Tables}
\crefname{table}{Tab.}{Tabs.}

\begin{document}
\title{On the Relevance of Temporal Features for Medical Ultrasound Video Recognition}
\titlerunning{Temporal Features for Ultrasound}
%
\author{D. Hudson Smith\inst{1}\orcidID{0000-0003-3041-4602} \and
John Paul Lineberger\inst{1}\orcidID{0000-0002-0267-9999} \and
George H. Baker\inst{2}\orcidID{0000-0001-9880-9665}}
\authorrunning{D. Smith et al.}
%
\institute{Clemson University, Clemson SC 29634, USA
\email{\{dane2,jplineb\}@clemson.edu} \and
Medical University of South Carolina, Charleston SC 29425, USA
\email{baker@musc.edu}}
\maketitle              
\begin{abstract}
  Many medical ultrasound video recognition tasks involve identifying key anatomical features regardless of when they appear in the video suggesting that modeling such tasks may not benefit from temporal features. Correspondingly, model architectures that exclude temporal features may have better sample efficiency. We propose a novel multi-head attention architecture that incorporates these hypotheses as inductive priors to achieve better sample efficiency on common ultrasound tasks. We compare the performance of our architecture to an efficient 3D CNN video recognition model in two settings: one where we expect not to require temporal features and one where we do. In the former setting, our model outperforms the 3D CNN – especially when we artificially limit the training data. In the latter, the outcome reverses. These results suggest that expressive time-independent models may be more effective than state-of-the-art video recognition models for some common ultrasound tasks in the low-data regime. Code is available at \verb+https://github.com/MedAI-Clemson/pda_detection+.

\end{abstract}
\section{Introduction and related work}
\label{sec:intro}
Ultrasound (US) is one of the most common imaging techniques in medical practice, with applications to fetal imaging, cardiac imaging, sports medicine, and more. With the rise of US for routine clinical care, there is a growing interest in applying computer vision techniques to automate or enhance the analysis of US imagery \cite{liu2019deep}. Many US examinations involve the collection of video clips showing different anatomical regions. The medical imaging community is in the early stages of applying techniques from the video recognition community to US recognition tasks. These applications face several challenges arising from the nature of US as an imaging modality, differences between US imagery and natural imagery, and the lack of large representative datasets. To make matters worse, the collection of large medical datasets is often unethical or prohibitively costly. There is, therefore, a significant need for efficient methods that can produce high levels of performance using the minimum number of samples. In this work, we propose an efficient US video recognition architecture that takes advantage the nature of common US recognition tasks.

To design an efficient US recognition architecture, it is necessary to consider the space of US recognition tasks and evaluate the algorithmic structures needed to efficiently capture the semantics in those settings. We posit that many of these tasks amount to the identification of specific visual characteristics at key moments in the clip. The identification of the {\it standard plane} in fetal head US depends on recognizing key structures in fetal brain tissue \cite{chen2015automatic,pu2021automatic}; the quality assessment of FAST clips \cite{taye2022deep} relies on the ability to recognize that key organs and other structures have been visualized in the clip; view identification relies on recognizing orientation of the anatomical structures in relation to one another \cite{kornblith2022development,howard2020improving}; and the quantification of heart function requires measurement of ventricular volumes at two key moments in the cardiac cycle \cite{sofka2017fully}. Based on these observations, we propose a novel {\it US Video Network} (USVN) that treats frames as independent and unordered. USVN constructs expressive video representations by combining information from multiple frames using a novel multi-head attention mechanism. We demonstrate a setting in which USVN yields better performance and far better sample efficiency than a competing model that includes temporal features. We also demonstrate that, in a setting where temporal dependence is important, USVN lags behind the competing model. These contrasting outcomes demonstrate the importance of tailoring the model architecture to the structure of the US recognition task in data-constrained settings.

A large body of work has addressed video recognition tasks, including object tracking \cite{luo2021multiple}, temporal action localization \cite{xia2020survey}, captioning \cite{amirian2020automatic}, action recognition \cite{zhang2019comprehensive}, and many others. Driven by the availability of large human action datasets, the field of action recognition has focused on the need to capture expressive spatiotemporal features. This has led to the development of two-stream networks using optical flow \cite{simonyan2014two}, the use of 3D convolutional networks \cite{ji20123d,tran2018closer}, and, of course, the use of transformer-based architectures \cite{plizzari2021spatial,mazzia2022action}. Our main point of departure with these methods is the importance placed upon temporal features. We posit that temporal features are not relevant in some common US tasks and that excluding these features leads to better sample efficiency. To explore this idea, we assume temporal independence {\it a priori}, placing our problem formulation in the format of a Multi-instance Learning (MIL) task. 

Multi-instance learning (MIL) describes the situation where labels apply to bags of instances rather than to individual instances. Instances within a bag are assumed to be unordered and, conditional on the bag label, independent from one another \cite{carbonneau2018multiple}. Under our assumption that all video frames can be treated independently, video recognition can be viewed as MIL where the bag is the video, and the instances are the frames. MIL has a long history of applications to video recognition that predates deep learning \cite{yang2005multiple,gu2008multi,stikic2009activity,ding2012horror}. In the classical formulation of MIL it is assumed that instances have unobserved labels, and the task is to extract these as latent variables and aggregate them to predict the bag-level label. In their paper {\it Attention-based deep multiple instance learning} Ilse, Tomczak, and Welling \cite{ilse2018attention} depart from this classical perspective by aggregating embeddings rather than instance labels. We take a similar approach. Unlike their work, however, we use multiple attention heads focused on different subspaces of the image-level embeddings, with their work as a special case of ours. To our knowledge, we are the first to introduce a MIL formalism using multiple attention heads in this way. 

There is growing interest in applying action recognition techniques to medical US video with applications to fetal \cite{chen2015automatic,pu2021automatic,rasheed2021automated}, abdominal \cite{kornblith2022development,taye2022deep}, and cardiac \cite{patra2017learning,howard2020improving,dezaki2017deep,sofka2017fully} US. Most existing applications make MIL assumptions but only apply a fixed pooling function to frame-level labels. Howard et al.~\cite{howard2020improving} apply a range of techniques, including average pooling, two-stream networks, and 3D convolutions to identifying cardiac views. They conclude that two-stream networks yield the best performance. The authors do not test any methods that adaptively pool frame information in a time-independent manner. Lei et al.~\cite{lei2022patent} specifically consider the detection of Patent Ductus Arteriosus (PDA). They make MIL assumptions by applying the video-level label to the individual frames and training a 2D CNN to estimate these noisy labels. Video-level labels are generated by applying a decision threshold to the frame-level predictions and then voting with equal weight across frames. Ouyang et al.~\cite{ouyang2020video} use 3D convolutions, specifically the R(2+1)D architecture \cite{tran2018closer}, to predict ejection fraction from cardiac US obtaining human-level performance. They do not assess the performance of any time-independent methods. Among these examples, we see a divide between methods that have no ability to adaptively weight different frames and those that can express arbitrary spatiotemporal features. We fill this gap by proposing a time-independent method that adaptively pools information from different moments in time.


\label{sec:method}
\begin{figure}[t]
  \centering
   \includegraphics[width=0.6\linewidth]{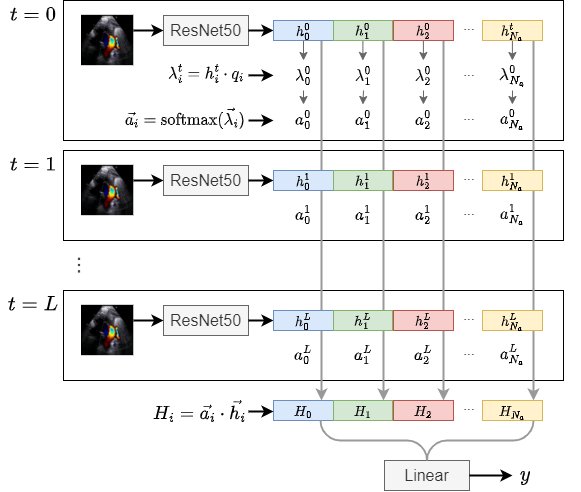}

   \caption{Proposed video-recognition architecture. Frame representations from ResNet50 are partitioned into $N_\mathrm{a}$ equal-sized vectors, $h_i^t$, represented by the colored boxes at each time step. These are compared by dot product with global query vectors $q_i$ to compute attention weights $a_i^t$. The video-level representation, $H_i$, is the attention-weighted sum of the partitions across frames. $y$ is the video-level prediction.}
   \label{fig:arch}
\end{figure}

\section{Proposed Method}
\subsection{USVN}
\label{sec:usvn}

\paragraph{Architecture.} Our video recognition architecture, shown in \cref{fig:arch}, pools information across frames using a multi-head attention mechanism. Like the attention mechanism in the transformer architecture \cite{vaswani2017attention}, we compute attentions over subspaces of the frame-level representations. We hypothesize that US video recognition requires the detection of distinct visual features that may appear at different points of time in the video. The individual attention heads can function as detectors of these features. Unlike ordinary multi-head attention, the subspaces are not compared with other frames in the sequence but with a set of global query vectors inferred during training. The use of global query vectors arises from our inductive prior that the recognition task amounts to locating key pieces of information at any point in the sequence, and the inferred query vectors are representations of that key information.

Frames are first embedded into $2048$-dimensional vectors using a CNN encoder. This encoder is initialized via ImageNet pretraining and fine-tuned during training. Rather than learn $N_a$ projections from scratch for the attention weighting, we simply partition the frame representations into $N_a$ vectors $h_i^t$ each of size $d_a = 2048/N_a$ and rely on the final convolutional layers of the CNN to adapt. We then compute the un-normalized attention scores via dot product with the global query vectors: $\lambda_i^t = h_i^t \cdot q_i$. The resulting scores are normalized resulting in $N_a$ attention vectors, $\vec{a_i} = \mathrm{softmax}(\vec{\lambda_i})$, where the arrow notation represents vectorization in time. The video-level representation from the $i^\mathrm{th}$ head is then simply $H_i = \vec{a_i}\cdot\vec{h_i}$, and the full video representation is the concatenation $H = \mathrm{concat}([H_1, H_2,\ldots, H_{N_a}])$. The video-level prediction can then be computed using a shallow fully-connected network, $y=f(H)$.

\paragraph{Augmentation by frame sampling.} Because USVN treats all frames independently, it is not necessary to use contiguous spans of frames during training. Instead, we randomly sample fixed-size sets of frames from each video. This can have a regularizing effect by using novel frames for each training epoch. During evaluation we use all video frames. We accommodate the varying numbers of frames in each video by zero padding and masked attention.

\paragraph{Model interpretability.} We identify prototype frames for each attention head. These prototypes produce embedding subspace vectors $h^t_i$ that are closely aligned with the corresponding query vector $q_i$. These prototype images can then be qualitatively evaluated by the clinical specialist (see Supplemental Material). 

\subsection{Benchmark implementations}

A simple and common approach for video recognition is to use fixed pooling functions to aggregate the frame-level representations across time, treating each element of the representation as a channel. We evaluate this approach using max and average pooling functions. Our attention-based method can implement average pooling by assigning equal weight to all frames for each attention head. Neglecting potential optimization challenges, this suggests that attention-based pooling should be at least as good as average pooling. On the other hand, our model can only approximate max pooling in the $N_a = 2048$ case by assigning very large, positive values to the single-element query vectors causing the attentions to become sharply concentrated at one time step. However, this solution pushes the softmax over time into regions with very small gradients. We conclude that max pooling can learn video representations that cannot be expressed by USVN (and vice versa).

R(2+1)D is a 3D CNN video recognition architecture that decomposes the spatial and temporal convolution into two successive steps\cite{tran2018closer}. First, a 2D convolution is applied over space then a 1D convolution is applied over time. Compared to its 3D ResNet counterparts on Sports-1M and Kinetics datasets, R(2+1)D is a very capable model that can learn complex features while having the same number of parameters in a more data-efficient way. We choose to benchmark against this architecture due to its efficiency and because this is the architecture used by Ouyang et al. to achieve human-level performance on the EchoNet-Dynamic US dataset \cite{ouyang2020video}. 


\section{Experimental Results}
\label{sec:results}

\subsection{Datasets}
\paragraph{Patent Ductus Arteriosus (PDA).} PDA is an opening between the aorta and pulmonary artery that, in severe cases, can cause heart failure shortly after birth. Ultrasound imaging is the primary diagnostic tool for detecting and characterizing PDA. Specifically, doppler US imaging can visualize the motion of the blood through the PDA opening. This motion appears as a characteristic blob of color in the region of the PDA. Physicians are trained to recognize the color and shape of the blob as well as where it appears in relation to other visible anatomy. Superficially, this recognition task makes no reference to the dynamics of the video. We therefore expect that temporal features are not required for accurate PDA recognition. For this dataset we train USVN to predict whether or not an image indicates the presence of PDA. The model output, $y$, is therefore a single number interpreted as the log-odds of PDA.

We retrospectively collected a set of 1,145 doppler US clips from 165 distinct examinations involving 66 distinct patients. Each clip was labeled to indicate the presence (661 clips) or absence (484 clips) of PDA. Patients were divided into training (44), validation (11), and test (11) sets with stratification on the presence of PDA. These sets contained 755, 118, and 272 videos, respectively. The large variation in the number of videos in the validation and test sets results from the fact that patients have a variable number of examinations ranging from 1 to 10.

\paragraph{EchoNet-Dynamic.} The Echonet Dynamic dataset consists of 10,030 apical-4 chamber echocardiograms downsampled to 112x112. Each study has clinical measurements: ejection fraction (EF), end systolic volume (ESV), and end diastolic volume (EDV). EF is commonly used to assess cardiac function and is computed from ESV and EDV as
\begin{equation}
  \mathrm{EF} = 1 - \mathrm{ESV} /\mathrm{EDV}.
  \label{eq:EF}
  \end{equation}
  The echocardiograms were obtained by registered sonographers and level 3 echocardiographers. For each of these videos, a masking and cropping transformation was performed to remove text and instrument information from the scanning area.

  For this dataset, we train USVN to predict ejection fraction. Rather than predict EF directly, we output a tuple of real numbers $(y_1,\,y_2)$ and insert them in place of ESV and EDV in \cref{eq:EF}. This choice is motivated by the knowledge that ESV and EDV are determined from different phases of the cardiac cycle. We speculate that decomposing EF into ESV and EDV effectively linearizes the estimation of EF as a function of the video representation $H$ with different attention heads responsible for estimating ESV and EDV. 

\subsection{Results}

\begin{table}[t]
\caption{Model performance comparison. EchoNet benefits from modeling temporal features; PDA does not. Performance is measured on the test set.}
\begin{tabularx}{6.5cm}{lcc}

  \toprule
                      & PDA       & EchoNet \\
  Model               & (ROC~AUC) & ($r^2$) \\
  \midrule
  R(2+1)D             & 0.816         & {\bf 0.822}     \\
  Average Pool        & 0.837         & 0.679           \\
  Max Pool            & 0.835         & 0.657           \\
  USVN (Ours)    & \bf{0.855}    & 0.765           \\
\bottomrule
\end{tabularx}
\label{tab:results}
\end{table}
\paragraph{Model performance.} \Cref{tab:results} summarizes the performance of USVN and our benchmark implementations on the PDA and EchoNet tasks. For PDA classification, we evaluate using the area under the ROC curve (ROC AUC). For EchoNet, we use the percent of variance explained ($r^2$). USVN results are based on $N_a=16$ and $N_a=128$ for PDA and EchoNet, respectively, based on a hyperparameter search (see Supplemental Material). For the PDA dataset, we expected that temporal features are not beneficial and, indeed, we see that R(2+1)D performs worse than all other methods, likely due to the unneeded capacity in the temporal convolutions and the relatively small size of the PDA dataset. USVN leads to a small benefit over average and max pooling for this task. The EchoNet task does benefit from modeling temporal features as indicated by R(2+1)D obtaining the highest score. However, USVN significantly outperforms the fixed pooling methods and is surprisingly close to R(2+1)D. This suggests that temporal features play a relatively small part in explaining the variability in the EchoNet dataset.

\begin{figure}[t]
  \begin{center}
    \includegraphics[width=0.48\textwidth]{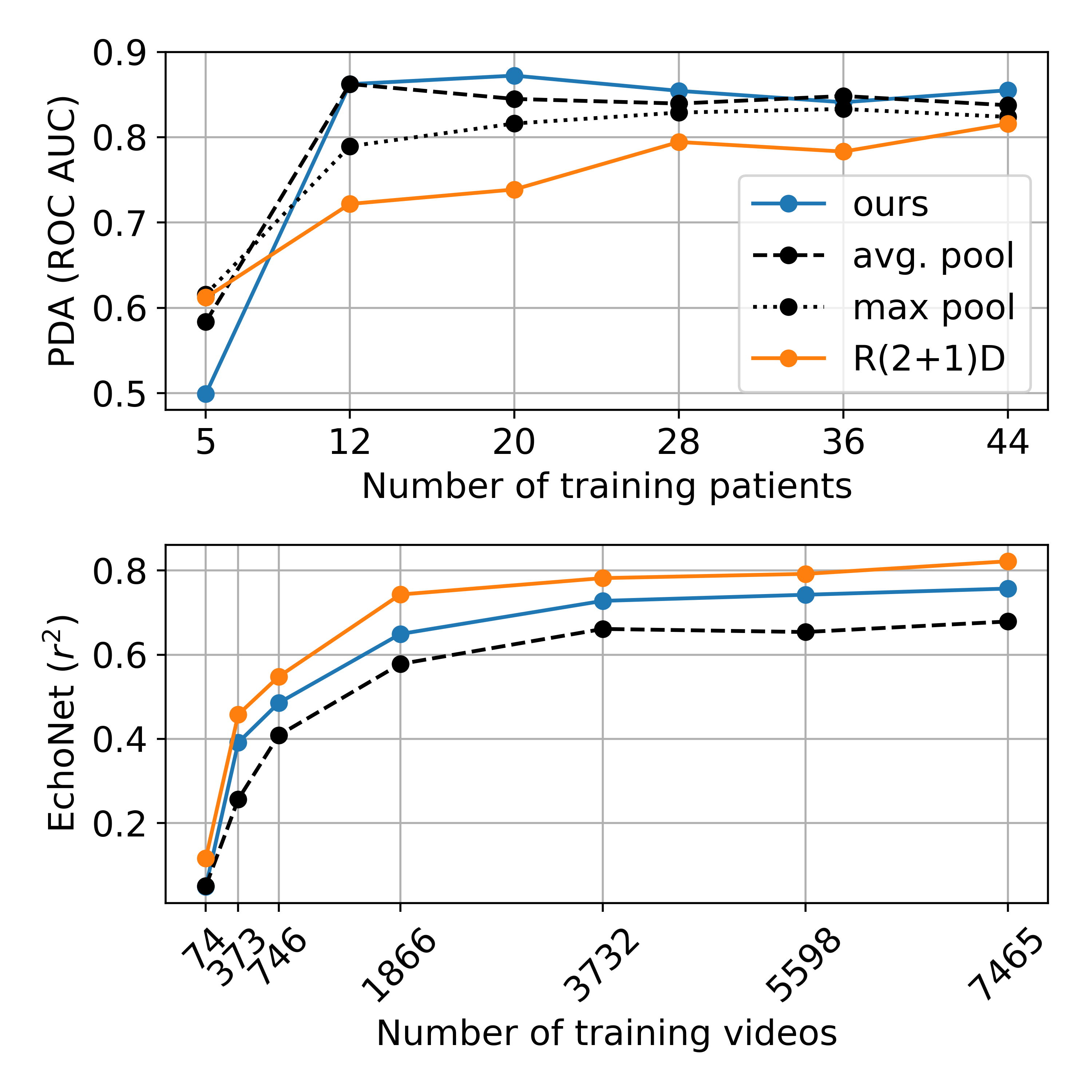}
  \end{center}
   \caption{Dependence on number of patients in training set for PDA classification ({\it top}) and EchoNet ejection fraction prediction ({\it bottom}). For PDA, we show patients, rather than videos along the x-axis due to the non-independence of videos from the same patient. For EchoNet, we omit the ``max pool'' variant because it failed to obtain positive $r^2$ values for several points along the x-axis. Performance is measured on the test set.}
   \label{fig:data_size}
 \end{figure}

 \paragraph{Sample efficiency.} In \cref{fig:data_size} we evaluate the sample efficiency of USVN by artificially limiting the amount of training data. In the case of PDA, we downsample the number of patients because videos from a single patient are correlated with one another. For EchoNet, we downsample the number of videos. In both cases, we use the full validation and test sets to better isolate variation due to limited training data from variation due to model selection and evaluation.

 For PDA, R(2+1)D underperforms the time-independent methods, and the gap is larger for smaller numbers of training patients (see \cref{fig:data_size}, top panel). Surprisingly, USVN and average pooling have very similar performance across samples and saturate for a small subset of the available patients.  R(2+1)D needs all available patients to approach a similar level of performance. This result aligns with our expectation that the inductive prior of time independence can yield sample efficiency benefits when applied to the appropriate task.

R(2+1)D outperforms the time-independent models across all samples for the EchoNet task (see \cref{fig:data_size}, bottom panel). Despite being a much simpler architecture than R(2+1)D and approaching similar levels of performance, USVN does not exhibit any sample efficiency benefits in the low-data regime for the EchoNet task. Solving the EchoNet task with spatial features alone may require more adaptation of the pretrained encoder than is required when solving with temporal features. For instance, it may be possible through extensive adaptation of the encoder network to recognize the visual characteristics associated with the end of diastole. However, the end of diastole may also manifest as, for example, an extremum in time of some visual characteristic. A model with access to temporal features such as R(2+1)D may be able to capture such an extremum with relatively little adaptation of the pretrained network.

 \subsection{Implementation details}
For the fixed pooling methods and USVN, we use an ImageNet-pretrained ResNet50 image encoder provided through the \verb+timm+ library \cite{rw2019timm}. We train using the \verb+timm+ implementation of the AdamP optimizer \cite{heo2020adamp} with $\beta_{1,\,2} = 0.9,\,0.999$, weight decay of 0.001, batch size of 20 clips, and initial learning rates of $3\cdot10^{-5}$ and $0.001$ for PDA and EchoNet, respectively. We sample 32 frames per clip during training. We reduce the learning rate by a factor of 10 after 3 epochs with no improvement of the validation loss, and we terminate training after ten consecutive epochs of no improvement. We use 50\% dropout on the inputs to the linear layer for each dataset. 

To reproduce the results of R(2+1)D on Echonet Dynamic Dataset by Ouyang et al.~\cite{ouyang2020video}, we cloned their github repo and re-ran their experiments with their best found hyperparameters. Our training runs show similar, if not better, results than stated in the original work. To adapt the model for PDA classification, we modified their data loader, training script, and the R(2+1)D model to allow PDA images. We also removed the manual bias term initialization, left over from predicting ejection fraction on the fully connected linear layer, and initialize it randomly instead. Finally, we replaced MSE loss with binary cross entropy with logits in the training loop. Every run was done for 45 epochs with a batch size of 20 for Echonet Dynamic dataset and 10 for PDA dataset. Model saving occurred for every epoch that showed improvement to the validation loss.

 \section{Conclusions and Discussion}

 The field of video recognition has been driven by large human action recognition datasets. Unlike videos of human actions, the accurate recognition of medical ultrasound images often only requires identifying key pieces of information at any point in the video and does not make reference to the sequence of events. The contrast between results for the PDA task (where USVN excels) and the EchoNet task (where USVN suffers) demonstrates the importance of tailoring the model architecture to the task at hand in data-constrained settings. Our results suggest that models developed for human action recognition are not optimal in some practical scenarios involving medical ultrasound and that models that assume temporal independence have better sample efficiency. We introduce an architecture, USVN, that is tailored to the medical ultrasound context and demonstrate a situation where the inductive prior of time independence leads to significant sample efficiency benefits. We also present a situation where temporal features are relevant and show that, even for very small datasets, USVN produces no efficiency benefits. Practitioners of deep learning who work with medical ultrasound in the low-data regime should take care to match the architecture choice to the nature of the recognition task.

 \section*{Acknowledgement}
 We thank Clemson University for their generous allotment of compute time on the Palmetto Cluster.

%
%
%
\bibliographystyle{splncs04}
\bibliography{bib}

\newpage
\appendix
\section{Selecting the number of attention heads.} 
 
\begin{figure}[h]
  \centering
  \includegraphics[width=0.45\linewidth]{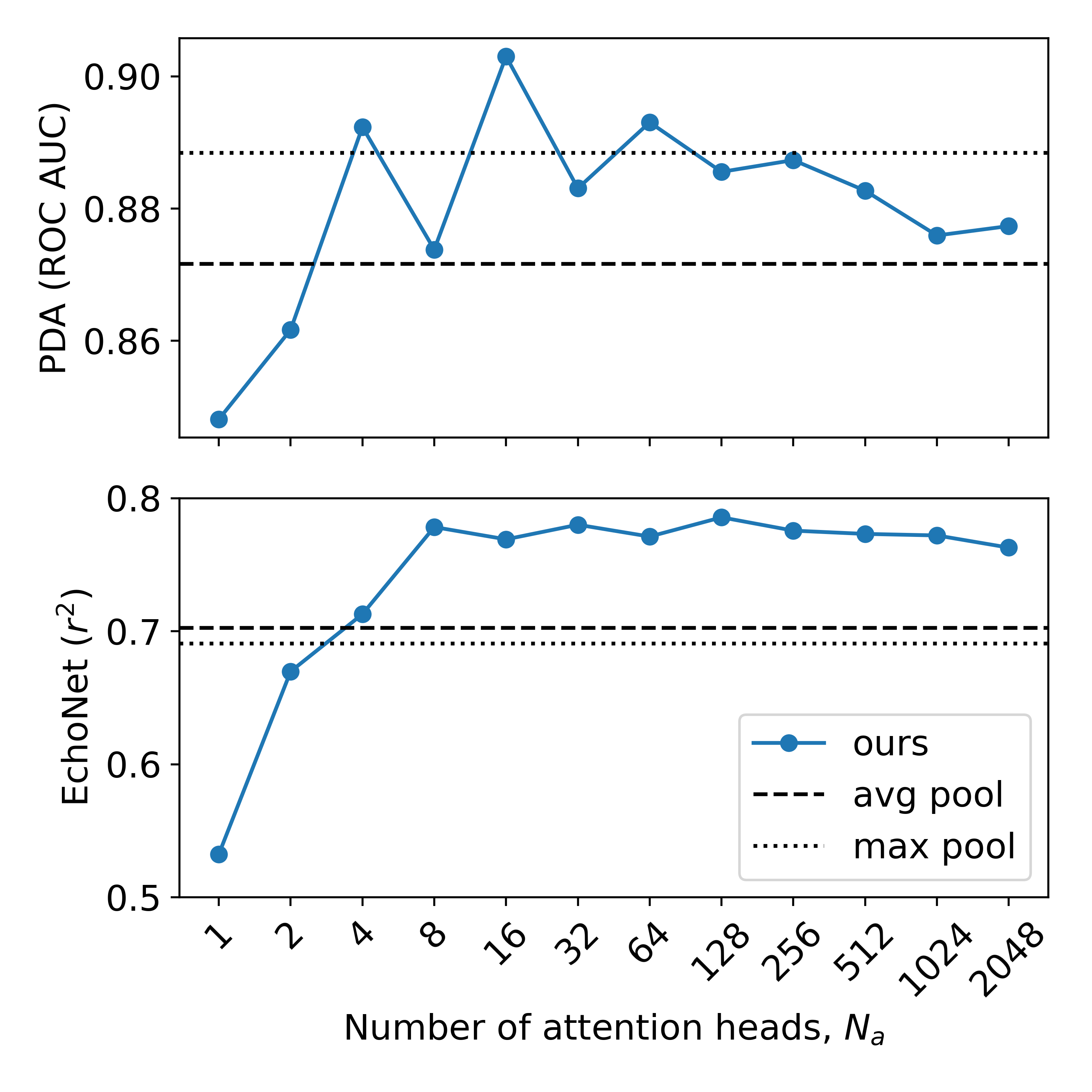}

   \caption{Dependence on number of attention heads for PDA classification ({\it top}) and EchoNet ejection fraction prediction ({\it bottom}). Performance is measured on the validation set.}
   \label{fig:num_heads}
 \end{figure}

 Each attention head can represent distinct visual characteristics important for the video recognition task. \Cref{fig:num_heads} shows the sensitivity of USVN's performance to the number of attention heads, $N_a$. For both tasks, several heads are better than one. For EchoNet, the performance remains relatively constant after 8 attention heads. We expect that the variation in the PDA results are due to the small size of the PDA dataset. Based on these results, we select $N_a=16$ for all other PDA experiments and $N_a=128$ for EchoNet.

 For small values of $N_a$, USVN underperforms average and max pooling for both tasks. This is surprising since USVN can emulate average pooling by assigning equal weight to all input frames. Our optimization procedure may perform poorly for small values of $N_a$. 

\section{Qualitative analysis.} 
 \begin{figure*}[t]
   \centering
   \includegraphics[width=1.0\linewidth]{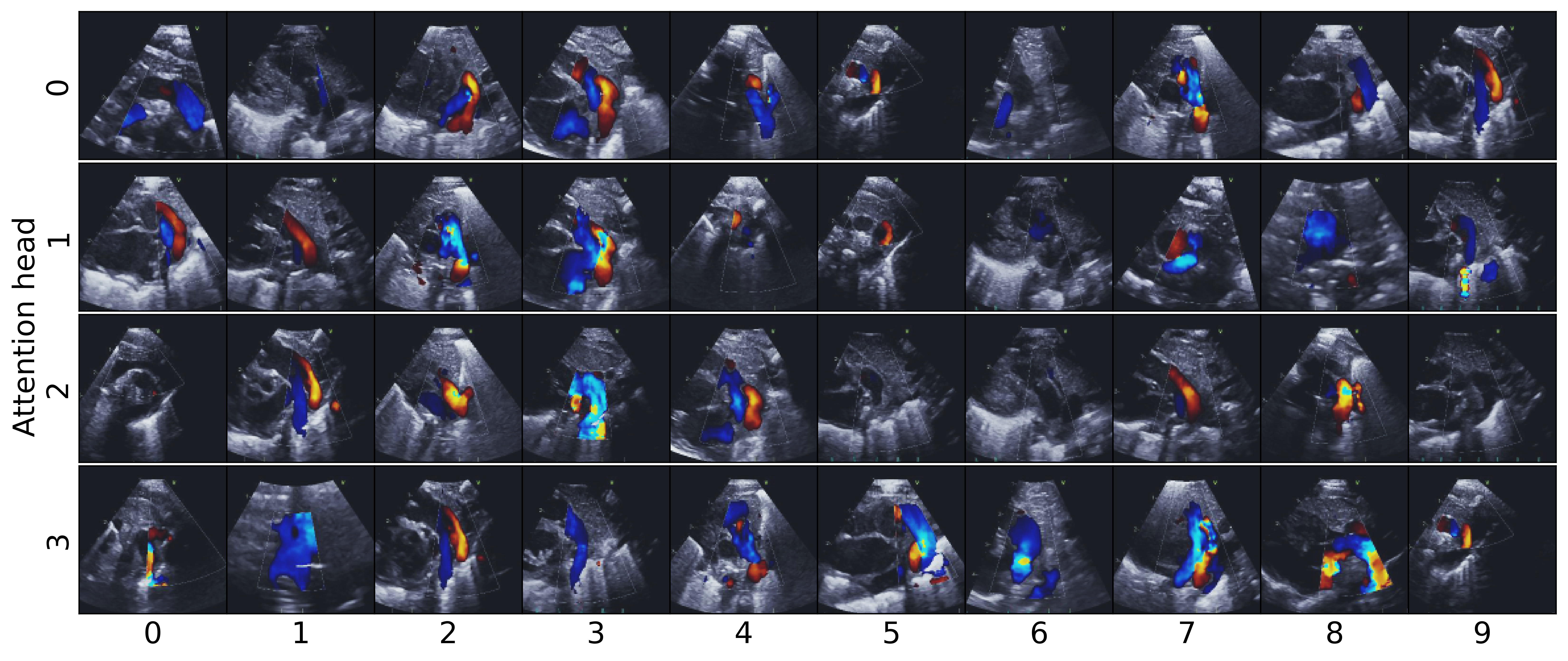}
   \includegraphics[width=1.0\linewidth]{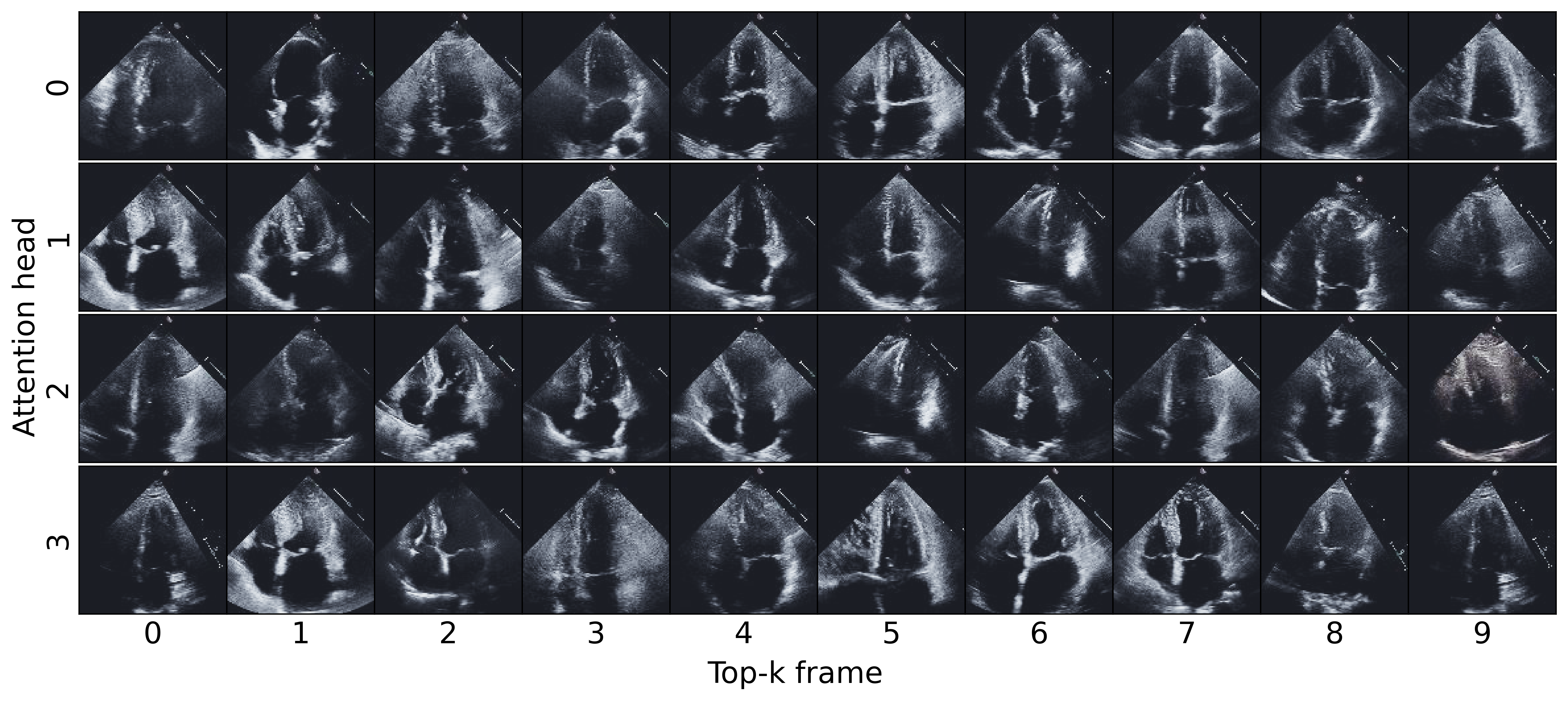}

   \caption{Examples of high-attention frames from the PDA ({\it top}) and EchoNet ({\it bottom}) datasets. Results are shown for the four attention heads with lowest average entropy in the time domain. In other words, these attention heads tend to focus on a small number of frames in each video. For each head, the 10 highest-attention frames were selected from a batch of 80 validation set videos.}
   \label{fig:attention_head_examples}
 \end{figure*}
We inspect USVN by finding video frames with high attention weights for several attention heads (see \cref{fig:attention_head_examples}). We identify the four heads that produce the lowest-entropy attention weights across time on average. Conceptually, heads with low entropy focus on a smaller number of frames in each video than heads with high entropy. We choose to inspect low-entropy heads on the assumption that these would be associated with more distinct visual characteristics. For each head, we identify the top-10 highest-attention frames across a batch of 80 videos.

From the clinician’s perspective, separate attention heads appear to focus on particular aspects of the pertinent anatomy and physiology represented by the Doppler color flow and the 2-D grayscale. For example, in the top ten highest-attention frames from attention head 0 (\cref{fig:attention_head_examples}, top), many frames demonstrate the red flow through the PDA while the remaining frames demonstrate an absence of flow in the PDA. This is intuitively similar to how a cardiologist would focus on the images. Additionally, in the top ten highest attention frames from attention head 2 we again see a number of frames with red flow through the PDA. Interestingly, we also see frames with no color doppler flow at all, suggesting focus on anatomic structures represented in the 2-D grayscale data rather than the color flow data. This is also intuitively similar to how an interpreting clinician would focus attention on the 2-D information to confirm the color flow they see is within the correct anatomic structure.

\section{Reproducibility Notes:}
\begin{itemize}
\item Study was conducted under IRB project number \verb+Pro00118191+ granted by the Institutional Review Board at the Medical University of South Carolina, Office of Research Integrity. 
\item USVN was developed in the Python programming language (version 3.9.12) using Pytorch (version 1.12.1).
\item Other python libraries: \verb+pandas, numpy, scipy, scikit-learn, matplotlib,+ \verb+seaborn, opencv-python, jupyter, jupyterlab, timm+
\item Models were trained using Clemson University's Palmetto Cluster using a single A100 GPU. 
\end{itemize}
\end{document}